\documentclass[acmtog, authorversion]{acmart}

\usepackage{tikz}
\usepackage{comment}
\usepackage{amsmath} 
\usepackage{color}
\usepackage{booktabs, multirow}
\usepackage{subfig}
\usepackage{gensymb}
\usepackage{tabularray}
\AtBeginDocument{%
  \providecommand\BibTeX{{%
    \normalfont B\kern-0.5em{\scshape i\kern-0.25em b}\kern-0.8em\TeX}}}

\setcopyright{acmcopyright}
\copyrightyear{2023}
\acmYear{2023}
\acmDOI{XXXXXXX.XXXXXXX}

\DeclareUnicodeCharacter{223C}{\textvisiblespace}
\begin{document}

\title{Efficient and Low-Footprint Object Classification using Spatial Contrast}

\author{Matthew Belding}
\email{mtb60@pitt.edu}
\affiliation{%
  \institution{Electrical and Computer Engineering, University of Pittsburgh}
  \city{Pittsburgh}
  \state{Pennsylvania}
  \country{USA}
  \postcode{15261-3323}
}
\author{Daniel C. Stumpp}
\email{daniel.stumpp@pitt.edu}
\affiliation{%
  \institution{NSF-SHREC Center, Electrical and Computer Engineering, University of Pittsburgh}
  \city{Pittsburgh}
  \state{Pennsylvania}
  \country{USA}
  \postcode{15261-3323}
}
\author{Rajkumar Kubendran}
\email{rajkumar.ece@pitt.edu}
\affiliation{%
  \institution{Electrical and Computer Engineering, University of Pittsburgh}
  \city{Pittsburgh}
  \state{Pennsylvania}
  \country{USA}
  \postcode{15261-3323}
}

\renewcommand{\shortauthors}{Belding, et al.}

\begin{abstract}
  Event-based vision sensors traditionally compute temporal contrast that offers potential for low-power and low-latency sensing and computing. In this research, an alternative paradigm for event-based sensors using localized spatial contrast (SC) under two different thresholding techniques, relative and absolute, is investigated. Given the slow maturity of spatial contrast in comparison to temporal-based sensors, a theoretical simulated output of such a hardware sensor is explored. Furthermore, we evaluate traffic sign classification using the German Traffic Sign dataset(GTSRB) with well-known Deep Neural Networks (DNNs). This study shows that spatial contrast can effectively capture salient image features needed for classification using a Binarized DNN with significant reduction in input data usage (at least 12×) and memory resources (17.5×), compared to high precision RGB images and DNN, with only a small loss ($\sim$2\%) in macro F1-score. Binarized MicronNet achieves an F1-score of 94.4\% using spatial contrast, compared to only 56.3\% when using RGB input images. Thus, SC offers great promise for deployment in power and resource constrained edge computing environments.
\end{abstract}




\maketitle

\section{Introduction}
\label{sec:introduction}
Artificial Intelligence (AI) has the potential to transform our lives through its near ubiquity in future technologies. However, with the tremendous scaling of AI platforms, which require ever-growing datasets, compute resources, and energy bills over the model lifecycle, carbon footprints of AI will grow exponentially over the next decade \cite{AIcarbon2019}. 
Researchers identify two key opportunities for improving energy efficiency and reducing $CO_{2e}$ emissions \cite{Google2021}: i) Large but sparsely activated DNNs ii) Cloud-vs-edge computing for machine learning (ML) workload scheduling. 

Both opportunities can be addressed by neuromorphic systems, which provide a scalable, hierarchical, \textit{sparse event-driven} computing architecture inspired by retina and cortical structures in the brain for efficient \textit{local information processing on-the-edge}. The field of neuromorphic engineering has grown rapidly in recent years, specifically in event-based vision. Asynchronous event-based vision sensors provide advantages such as high dynamic range and reduced data output when compared to traditional frame-based sensors that rely on synchronized capture of frames \cite{gallego2020event}. While most research in neuromorphic vision has focused on the use of temporal contrast (TC) event-based vision sensors, there are other neuromorphic vision sensor variations such as local spatial contrast sensors.  

In this paper, we explore the use of spatial contrast (SC) sensors coupled with a variety of lightweight machine learning networks for efficient traffic sign classification in extremely resource constrained environments. Object detection\footnote {See Supplemental findings for investigation of object detection using SC sensors at \color{blue}\url{https://github.com/danielstumpp/neuro-spatial-contrast.git}.} and classification is a fundamental challenge for autonomous vehicles. With the evolution of autonomous driving, starting from advanced driver assist features to eventually full autonomous capability, the amount of data collected by vehicle sensors has increased tremendously. It is estimated that future connected and autonomous vehicles may produce up to 5 TB of data per hour \cite{li2019vehicle}. This necessitates high performance hardware to process the high resolution, high throughput data with sufficiently low latency ($<$1ms), for real-time critical decision making. Using images generated from SC sensors has the potential to enable low latency, energy efficient detection and classification with reduced memory footprint. This paper analyzes the use of SC images for traffic sign classification and discusses the performance impact of various thresholding methods proposed. Additionally, we compare classification F1-score for a variety of networks when using SC images as opposed to traditional RGB input images. 

\section{Background}
\label{sec:background-sota}
Neuromorphic engineering pursues the hardware/algorithm co-design of electronic systems emulating function and structural organization of biological neural systems. Neuromorphic systems often embody similar physical principles found in biology and are optimized for extreme energy efficiency. Though this field started three decades ago, it has gained widespread recognition recently with the commercialization of neuromorphic event-based cameras called Dynamic Vision Sensors (DVS) \cite{gallego2020event}. DVS cameras capture temporal contrast (TC) in the individual pixels as ON/OFF events indicated by a increase/decrease in light intensity over a period of time. The events generated are asynchronously transmitted for further processing downstream. TC detection is great for capturing motion in the visual scene, where static objects produce limited events if any, depending on luminance. Hence for object detection and classification, DVS cameras are not useful unless there is motion. However, if spatial contrast (SC) can be captured instead of temporal contrast, it is possible to use such a neuromorphic image sensor to detect edges and object contours in a visual scene with minimal data throughput.

Spatial Contrast in a grayscale image is the difference between a pixel's intensity value and it's surrounding pixels. Previous work showed the feasibility of event-based pixel level contrast sensors \cite{sc-sensor,scamp2018,kubendran2021}. These sensors can provide ultra-low power imaging solutions and a drastic reduction in the output data. SC sensors need to improve further in resolution and availability to be practical for adoption. The results of this study will shed light on the potential impact of those devices when they reach maturity. The proposed techniques can be used as a post-processing step for traditional cameras to extract SC information and use it exclusively for pattern recognition applications. Even with existing camera sensors, the proposed methods can lead to significant savings in data usage, overhead latency in pre-processing images and memory footprint. 

\section{Methods}
\label{sec:methods}

\subsection{Dataset}
\label{sec:methods-dataset}

The German Traffic Sign Recognition Benchmark (GTSRB) dataset contains 51,840 1360$\times$1024 resolution images with German road signs for 43 unique classes. Training images are grouped by tracks, each of which contain 30 images for one single traffic sign. A fixed number of images were kept per track to increase diversity and avoid potential overfitting due to an abundance of nearly identical images\cite{Houben-IJCNN-2013}. The extracted signs from these images can vary in illumination as well as resolution. The size of the signs vary between 15$\times$15 $px$ and 250$\times$250 $px$, and contain a 10\% pixel margin.

GTSRB applies stratified sampling to preserve class distribution, splitting the dataset at random into three subsets while considering class and track. The final split is a training and test set that has a 75/25 split. Temporal information is preserved in the training set but not in the test set. To alleviate the imbalance of data amongst classes in the training set, we opted to upsample classes artificially to 400 samples using Tensorflow's $ImageDataGenerator$ \cite{tensorflow2015-whitepaper}. Augmentation methods applied include rotation, width shift, height shift, and zoom. Rotational augmentation consisted of a 20$^{\circ}$ range while the width, height and zoom methods use up to 20\% of the image as their respective maximum range. Any classes that contained larger than 400 samples were randomly undersampled to 400 to create a uniform class distribution for training. Validation and test sets maintained their original class distributions to remain as representative as possible to the real world. Due to the sample imbalance amongst classes in the test set, macro F1-score is chosen as the primary measure of performance in this study. This is further explained in Section~\ref{sec:experiments-and-results} and is hereafter referred to as simply F1-score.

\subsection{Thresholding Techniques}
\label{sec:thresh-tech}

We investigate two SC thresholding techniques in this study: absolute and relative. A  comparative performance analysis between absolute and relative is made. Threshold technique impact on classification F1-score is further discussed with results in Section~\ref{sec:experiments-and-results}.

\subsubsection{Absolute}
\label{sec:thresh-abs}
The simplest thresholding technique is absolute thresholding. A $3\times3$ averaging kernel is first applied to the grayscale input image to generate the neighbor average image, where each pixel's intensity is equal to the average intensity of the 8 neighboring pixels in the original image. This average image is then subtracted from the original grayscale image, resulting in the SC image. The event at each pixel is then determined using \eqref{eq:abs-thresh} where $I_{abs}$, $I_{sc}$, and $\rho$ represent the absolute thresholding event polarity, SC intensity, and the predefined absolute threshold respectively. The event polarity constitutes three possible values only, +1/-1/0, with 1 representing an `ON' event, -1 representing an `OFF' event, and 0 for `NO' event.

\begin{equation}
    I_{abs} = 
    \begin{cases}
        1 & I_{sc}\geq \rho \\
        -1 & I_{sc}\leq -\rho \\
        0 & otherwise
    \end{cases}
    ~~~~0 \leq \rho \leq 1
    \label{eq:abs-thresh}
\end{equation}

\begin{figure}
    \centering
    \includegraphics[scale=.48]{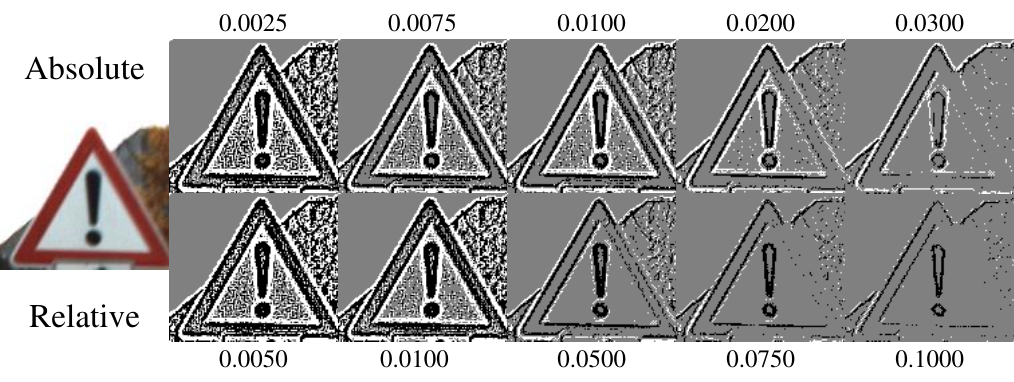}
    \caption{Examples of spatial contrast image representations with varying absolute and relative thresholds. Absolute thresholds are listed across the top, corresponding to the top row of images. Relative thresholds are listed across the bottom corresponding to the bottom row of images.}
    \label{fig:sc-examples}
\end{figure}

As seen in Fig.~\ref{fig:sc-examples}, as the threshold is increased, the number of events detected decreases, but the proportion of ON to OFF events remains relatively constant. The absolute thresholding technique has the advantage of being simple to implement, but with the disadvantage of sensitivity to varying image lighting depending on the regional illumination of the scene. When different illumination is present in an image, the ideal threshold to extract features in one part of the image may result in too few or too many events in another part of the image.

\subsubsection{Relative}
\label{sec:thresh-rel}
In an effort to address the issues inherent with absolute thresholding, we introduce a relative thresholding technique. The relative thresholding technique uses the same method of extracting the SC as discussed above. However, an additional normalization is added and the threshold range is modified. This modified equation is shown in \eqref{eq:rel-thresh} where $\beta$ is the relative threshold, $I_{rel}$ is the relative pixel event intensity, and $I_{gray}$ is the grayscale image intensity of the pixel.

\begin{equation}
    I_{rel} = 
    \begin{cases}
        1 & \frac{I_{sc}}{I_{gray}}  \geq \beta \\
        -1 & \frac{I_{sc}}{I_{gray}} \leq -\beta,I_{gray}\neq 0 \\
        0 & otherwise
    \end{cases}
    ~~~~0 \leq \beta < \infty
    \label{eq:rel-thresh}
\end{equation}

As the relative threshold is increased the number of events declines as seen with the absolute threshold. It is also important to note that as the threshold increases, the output is heavily biased to OFF events. This is because OFF events occur at pixels with low intensity, making the value of $\frac{I_{sc}}{I_{gray}}$ potentially very large. However, the maximum value of $\frac{I_{sc}}{I_{gray}}$ for ON events is one. This means that if the relative threshold were to be set $>$1, only OFF events would be seen. This behaviour can be seen in the bottom row of Fig.~\ref{fig:sc-examples}, where OFF events dominate at higher thresholds, but still allow for clear visual distinction between features.

The relative thresholding SC method provides more robustness to changing illumination throughout the image than absolute thresholding. It allows significant reduction of the event activity without loosing the important features in a scene. These gains are achieved with minimal increase in implementation complexity. 

\subsection{Network Selection}
\label{sec:network-selection}
Several architecture configurations were considered in this study. Due to their development for edge devices, MobileNetV2 and MicronNet were chosen for evaluation. MobileNetV2 introduced the inverted residual layer module with linear bottleneck to reduce the network's memory footprint \cite{sandler2019mobilenetv2}. MicronNet leverages macroarchitecture design and optimization to reduce size and the number of MAC (multiply-accumulate) operations \cite{wong2018micronnet}.

The base MobileNetV2 model contains nearly 300M MAC operations and 3.4M parameters. As this was designed for the ImageNet dataset, we reduce the starting baseline MobileNetV2 model using the width scaling option, $\alpha$, provided with the model. The scaling values tested were 0.35, 0.25, 0.15, and 0.05. Further parameter reductions were also tested at an $\alpha = 0.05$ by removing 2, 6, and then 10 residual bottleneck layers, respectively. 

MicronNet contains nearly half a million parameters using 10.5M MAC operations. As this was originally tested on the GTSRB dataset, it was suited as a reasonable architecture to test on the traffic sign classification task. This network was further reduced by binarizing the weights and activations of the original network. 

\section{Experiments and Results}
\label{sec:experiments-and-results}

The remainder of this paper explains the theoretical benefits of a SC sensor output for traffic sign classification. These benefits include a substantial data rate reduction, near-comparable classification performance to a traditional RGB sensor, and resiliency with parameter reduction and binarization.      

\subsection{Data Rate Reduction}
\label{sec:er-data-rate-reduction}
\begin{figure}[h]%
    \centering
    \subfloat[\centering]{{\includegraphics[scale = .55]{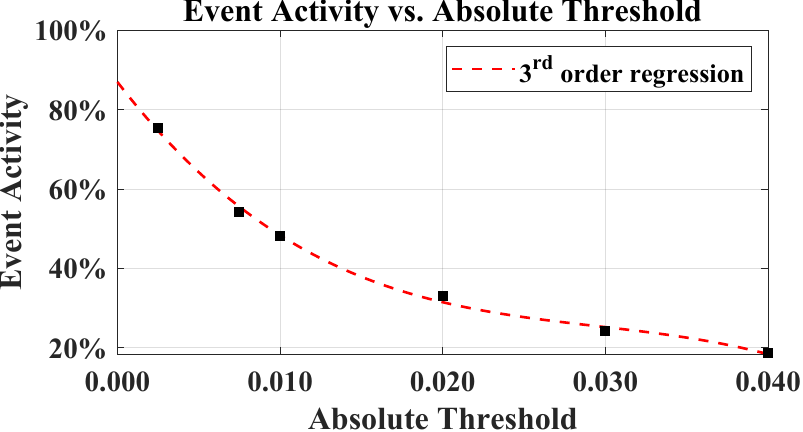} }}
    \qquad
    \subfloat[\centering]{{\includegraphics[scale = .55]{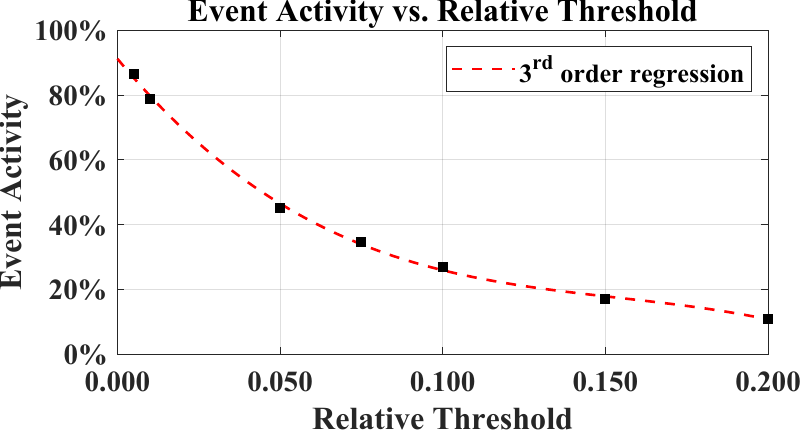} }}
    \caption{Events in image plotted against the (a) absolute threshold and (b) relative threshold applied.}
    \label{fig:events-vs-thresh}
\end{figure}

The use of a neuromorphic SC sensor enables the possibility of a significant reduction in the amount of data required to be transmitted and processed. The extent of this data rate reduction is impacted by the thresholding technique (absolute/relative), applied threshold, and the architecture of the sensor being used. Fig.~\ref{fig:events-vs-thresh} shows the event activity in percentage of active pixels plotted against the relative and absolute thresholds applied. The results in both cases are well fitted by a third order polynomial. As expected, increasing thresholds result in a deceasing level of activity in the image as less SC events are generated. 

\begin{figure}[h]%
    \centering
    \subfloat[\centering]{{\includegraphics[scale = .55]{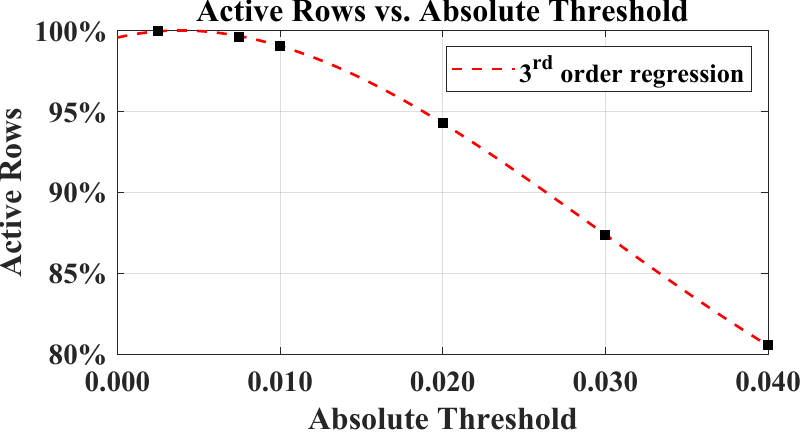} }}
    \qquad
    \subfloat[\centering]{{\includegraphics[scale = .55]{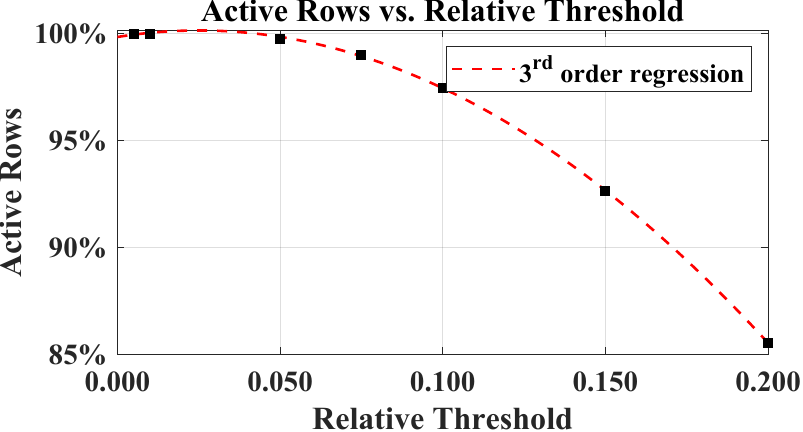} }}
    \caption{Active rows in image plotted against the (a) absolute threshold and (b) relative threshold applied.}
    \label{fig:active-rows-vs-thresh}
\end{figure}

While the event activity shown in Fig.~\ref{fig:events-vs-thresh} represents the minimum number of events required to be transmitted, the actual number of events that is sent from the sensor for further processing depends on the sensor hardware architecture. Only for fully asynchronous event detection and transmission, where any pixel can raise an event request at any time that can be acknowledged asynchronously, it is possible to transmit only the event activity detected. For a more practical approximation of the data transmission requirements, where synchronous frame scanning is used to readout rows and columns in the frame, a query based sensor \cite{kubendran2021} that only transmits rows with at least one active pixel is considered. Event activity represents the percentage of pixels that have an event in a given SC image. Active rows represents the percentage of rows in the image that have at least one active event in them. Fig.~\ref{fig:active-rows-vs-thresh} shows the average percentage of active row in the SC images when applied to the dataset. Upon observation, the number of active rows in the SC images decreases at a significantly reduced rate when compared to the event-rate results shown in Fig.~\ref{fig:events-vs-thresh}. This indicates that while some data rate reduction can be achieved from the sparsity of events, the majority of the reduction will come from the reduced size of pixel intensity values enabled by SC.

Generally, the data size of an image to be transmitted can be represented by the product of the image width ($w$), height ($h$), channels ($c$), bits per pixel channel ($b$), and the percent of active rows ($\alpha$) as shown in \eqref{eq:data}.

\begin{equation}
\centering
\label{eq:data}
    bits =  w \cdot h \cdot c \cdot b \cdot \alpha
\end{equation}

For an RGB image, there are three channels with eight bits per channel and all rows are considered active. SC images require only one channel, immediately providing a 3$\times$ reduction in data when compared to RGB images. Additionally each channel requires only two bits to represent the `ON', `OFF', and `NO' event states that a SC pixel can take. This provides a further 4$\times$ reduction, resulting in a 12$\times$ reduction in data compared to RGB images even when $\alpha = 1$ for the SC image. As the threshold is increased, the number of active rows is decreased causing additional reductions in the required data transmission sizes. This comes at a cost in F1-score shown in Section~\ref{sec:er-network-reduction}. 

\subsection{Network Reduction}
\label{sec:er-network-reduction}
Increasing the threshold of a technique can further reduce input data size but can incur a loss of relevant information. This is seen in an absolute threshold of 0.020, where an active row reduction to roughly 95\% results in a corresponding event activity of 33\%. This offers less information to a classification network and has potential to be detrimental to its performance. To evaluate the resiliency of SC in retaining its F1-score close to the RGB baseline, varying thresholds were tested for both contrast techniques against RGB for our baseline MobileNetV2 network ($\alpha = 0.35$). Macro F1-score captures a harmonic mean of precision and recall metrics and is therefore a better measure of incorrect classifications within our imbalanced test set. 

\begin{figure}
    \centering    \includegraphics[scale=.55]{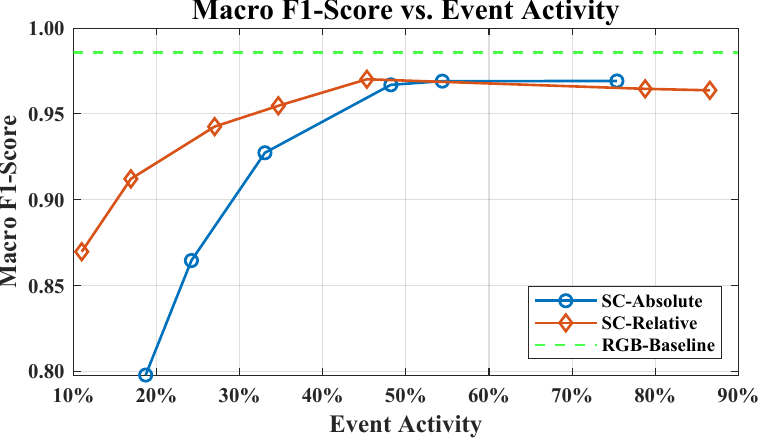}
    \caption{F1-score plotted with corresponding event activity in the image for absolute and relative thresholding along with the RGB baseline. Results shown are with the baseline network. }
    \label{fig:f1-vs-events}
\end{figure}

Fig.~\ref{fig:f1-vs-events} demonstrates not only SC's ability to stay within roughly 3\% of the baseline F1-score of 98.6\% with the need for only 30\% event activity, but also showcases its ability to maintain representative and valuable information that distinguishes one sign from another with more than 12$\times$ less data. This performance collapses below the 30\% event activity mark which can be attributed to SC's inability to maintain a clean representation of certain signs.

\begin{figure}[h]%
    \centering 
    \subfloat[\centering]{{\includegraphics[scale=.55]{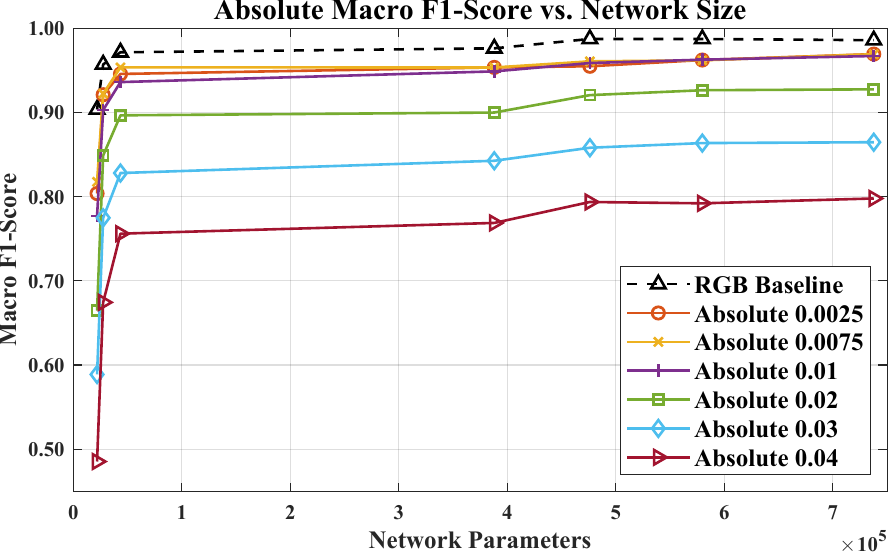} }}
    \qquad
    \subfloat[\centering]{{\includegraphics[scale=.55]{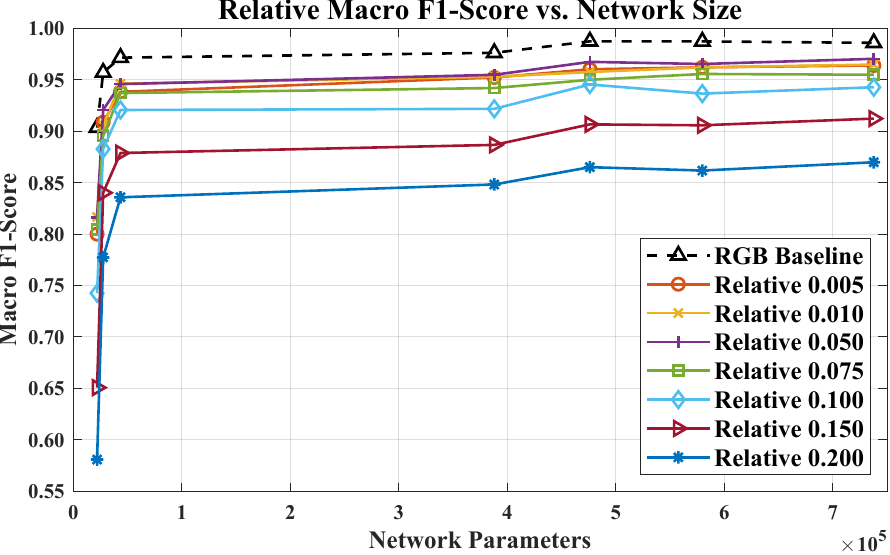} }}
    \caption{(a) F1-score plotted with corresponding network size for absolute thresholding along with the RGB baseline. (b) F1-score plotted with corresponding network size for relative thresholding along with the RGB baseline.}
    \label{fig:abs-f1-vs-network-params}
\end{figure}

To further identify benefits of SC, the network size of the baseline MobileNetV2 was reduced six times to measure performance across various thresholds. Fig.~\ref{fig:abs-f1-vs-network-params} represents the findings of such reductions. As expected, the three lowest thresholds for both methods outperform the rest of the thresholds by a fairly large margin of 3-5\% in F1-score. They also maintain within 2.5\% of the baseline's 97.1\% F1-score with a 17$\times$ reduction in parameters from 737,675 down to 43,403. The resulting reduced memory footprint is 170KB and roughly 21.3M MAC operations. When training parameters dip below 27,000, the classification accuracy begins to collapse for all thresholds including the RGB baseline.

\begin{figure}
    \centering
    \includegraphics[scale=.7]{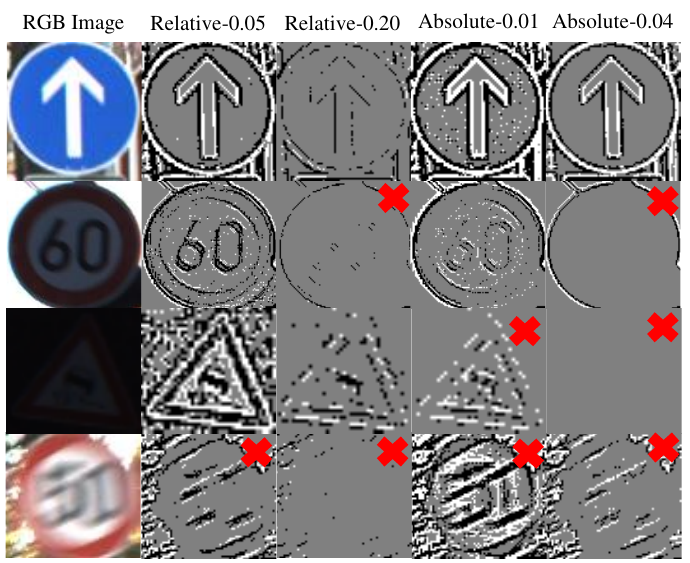}
    \caption{Sampling of test images with spatial contrast applied. Images marked with an `X' were incorrectly classified by the baseline MobileNetV2 network.}
    \label{fig:mobilenetv2-montage}
\end{figure}

In Fig.~\ref{fig:mobilenetv2-montage}, a subset of test images with different thresholding techniques and levels are shown. Relative-0.05 and Absolute-0.01 represent the best performing thresholds respectively, while Relative-0.20 and Absolute-0.04 represent the highest thresholds used. Images incorrectly classified by the baseline MobileNetV2 model are marked with an `X'. The signs shown represent a range of varying illumination and image quality. The first row shows the ideal case of a high-resolution, well-illuminated, high-contrast sign. All representations of this sign were correctly classified, and it can be visually observed that SC clearly captures the main features of the sign. The bias of the higher relative threshold towards `OFF' events should also be noted. The second row shows an image that is high resolution with respect to the overall dataset, but has lower SC due to lower illumination. In this case, both of the higher thresholds fail to capture the salient features of the sign and therefore their representations cannot be correctly classified. The third row shows an extremely low-light image that is barely visible. The relative thresholding does a much better job of extracting the visual information than absolute, which is reflected in both absolute representations being classified incorrectly. Shown in the last row is a case where the RGB image is correctly classified, but all SC representations fail. The blurring in the image prevents a quality SC image from being generated. While this represents a weakness of using SC as a  post-processing step, these types of errors would be mitigated by the use of a neuromorphic SC sensor with autonomous pixels that compute local SC at the sensor itself. The contrast events computed at the pixel level could then be queried with high temporal precision, using synchronous frame scanning.

\subsection{Network Binarization}
\label{sec:er-net-bin}
MicronNet represents a nearly 1.4$\times$ reduction in model size compared to the baseline MobileNetV2 and a 5.6$\times$ reduction in MAC operations. As shown in Table~\ref{table:binarization}, this architecture does not compromise accuracy at a significantly reduced computational cost and size, as it performs nearly identically to the baseline on both RGB and SC with scores of 98.3\% and 96.4\%, respectively. It also requires only about half the number of MAC operations as the most reduced version of MobileNetV2. To further investigate improving the efficiency of the network, the impact of binarizing the MicronNet architecture is studied.

In this study, the Larq Compute Engine is utilized for binarizing the MicronNet architecture \cite{bannink2021larq}. This binarization converted all weights and activations with the exception of the first and last layers of the network. The result is a near fully binarized MicronNet with a memory footprint of 115KB, reduced from 1.97MB. As seen in Table~\ref{table:binarization}, SC takes only a 2\% reduction in F1-score after binarization, while the RGB baseline F1-score is reduced by 42\%. This shows the ability of the SC representation to still be accurately classified with extreme reductions in model precision as well as model size. Furthermore it demonstrates the significance of using a binarized neural network on event-driven representations to further reduce computational cost, while achieving accuracy unattainable with traditional input images with the same network. Hence, coupling SC sensors with a binarized DNN is ideal for edge computing scenarios, since the data and memory requirements are significantly ($>$90\%) reduced while maintaining comparable performance in the F1-score.      

\begin{table}
\centering
\caption{Comparison of MobileNetV2 and MicronNet variants for relative threshold of 0.05. Results show SC significantly outperforms RGB baseline when using a binarized network. Binarized MicronNet is denoted bMicronNet. The best F1-Score for each network is indicated in bold.}
\begin{tblr}{
  row{even} = {c},
  row{1} = {c},
  row{3} = {c},
  cell{1}{1} = {r=2}{},
  cell{1}{2} = {r=2}{},
  cell{1}{3} = {r=2}{},
  cell{1}{4} = {c=2}{},
  cell{5}{1} = {c},
  cell{5}{2} = {c},
  cell{5}{3} = {c},
  cell{5}{4} = {c},
  hline{1,6} = {-}{0.08em},
  hline{2} = {4-5}{0.03em},
  hline{3} = {-}{0.05em},
}
\textbf{Network } & \textbf{Parameters } & {\textbf{Network}\\\textbf{Size (MB) }} & \textbf{F1-Score } &                          \\
                  &                      &                                         & \textbf{RGB}       & \textbf{SC}              \\
MobileNetV2       & 737675               & 2.81                                    & \textbf{0.986}     & 0.970                    \\
MicronNet         & 514491               & 1.97                                    & \textbf{0.983}     & 0.964                    \\
 bMicronNet       & 514491               & 0.115                                   & 0.563~             & \textbf{\textbf{0.944 }} 
\end{tblr}
\label{table:binarization}
\end{table}
\section{Conclusion}
\label{sec:conclusion}
The emergence of neuromorphic computing and the advent of event-based vision sensors has provided the opportunity to achieve low-power sensing and computation for tasks that have traditionally been computationally expensive and energy inefficient. In this paper, the use of spatial contrast images for efficient classification of traffic signs using different neural networks is investigated. Two methods of thresholding are presented, both of which are simple enough to be integrated into neuromorphic hardware or readily incorporated into traditional Computer Vision pipelines as a simple pre-processing step. 

It was demonstrated that the use of relative thresholding outperforms absolute thresholding due to robustness to variation in illumination across an image. Furthermore, we have shown that spatial contrast representations can be classified with an F1-score within 3\% of the RGB baseline by MobileNetV2, while requiring at least 12$\times$ less data. We have also demonstrated the significant performance advantage of spatial contrast image representations when using a binarized neural network for classification. The binarization of MicronNet resulted in a 17$\times$ reduction in the memory resources to store the weights of the network. In this study, we have further shown that, using a binarized version of MicronNet the spatial contrast images can be classified with an F1-score of up to 94.5\%, while the same signs represented by traditional RGB images are only classified with an F1-score of 56.3\%. This represents a tremendous advantage for spatial contrast images when using a binarized network. 

The ability to perform classification tasks with high accuracy using the combination of BNNs and spatial contrast data shows great promise for the deployment of event-based machine learning in power and resource constrained edge computing environments where current state-of-the-art techniques are highly inefficient. With tremendous savings in data and memory usage, the proposed method enables computing on-the-edge.

\begin{acks}
Daniel Stumpp was supported by NSF Center for Space, High-Performance, and Resilient Computing (SHREC) industry and agency members and by the IUCRC Program of the National Science Foundation under Grant No. CNS-1738783.
\end{acks}

\bibliographystyle{ACM-Reference-Format}
\bibliography{sample-base}


\begin{thebibliography}{12}


\ifx \showCODEN    \undefined \def \showCODEN     #1{\unskip}     \fi
\ifx \showDOI      \undefined \def \showDOI       #1{#1}\fi
\ifx \showISBNx    \undefined \def \showISBNx     #1{\unskip}     \fi
\ifx \showISBNxiii \undefined \def \showISBNxiii  #1{\unskip}     \fi
\ifx \showISSN     \undefined \def \showISSN      #1{\unskip}     \fi
\ifx \showLCCN     \undefined \def \showLCCN      #1{\unskip}     \fi
\ifx \shownote     \undefined \def \shownote      #1{#1}          \fi
\ifx \showarticletitle \undefined \def \showarticletitle #1{#1}   \fi
\ifx \showURL      \undefined \def \showURL       {\relax}        \fi
\providecommand\bibfield[2]{#2}
\providecommand\bibinfo[2]{#2}
\providecommand\natexlab[1]{#1}
\providecommand\showeprint[2][]{arXiv:#2}

\bibitem[Abadi et~al\mbox{.}(2015)]%
        {tensorflow2015-whitepaper}
\bibfield{author}{\bibinfo{person}{Mart\'{i}n Abadi}, \bibinfo{person}{Ashish
  Agarwal}, \bibinfo{person}{Paul Barham}, \bibinfo{person}{Eugene Brevdo},
  \bibinfo{person}{Zhifeng Chen}, \bibinfo{person}{Craig Citro},
  \bibinfo{person}{Greg~S. Corrado}, \bibinfo{person}{Andy Davis},
  \bibinfo{person}{Jeffrey Dean}, \bibinfo{person}{Matthieu Devin},
  \bibinfo{person}{Sanjay Ghemawat}, \bibinfo{person}{Ian Goodfellow},
  \bibinfo{person}{Andrew Harp}, \bibinfo{person}{Geoffrey Irving},
  \bibinfo{person}{Michael Isard}, \bibinfo{person}{Yangqing Jia},
  \bibinfo{person}{Rafal Jozefowicz}, \bibinfo{person}{Lukasz Kaiser},
  \bibinfo{person}{Manjunath Kudlur}, \bibinfo{person}{Josh Levenberg},
  \bibinfo{person}{Dandelion Man\'{e}}, \bibinfo{person}{Rajat Monga},
  \bibinfo{person}{Sherry Moore}, \bibinfo{person}{Derek Murray},
  \bibinfo{person}{Chris Olah}, \bibinfo{person}{Mike Schuster},
  \bibinfo{person}{Jonathon Shlens}, \bibinfo{person}{Benoit Steiner},
  \bibinfo{person}{Ilya Sutskever}, \bibinfo{person}{Kunal Talwar},
  \bibinfo{person}{Paul Tucker}, \bibinfo{person}{Vincent Vanhoucke},
  \bibinfo{person}{Vijay Vasudevan}, \bibinfo{person}{Fernanda Vi\'{e}gas},
  \bibinfo{person}{Oriol Vinyals}, \bibinfo{person}{Pete Warden},
  \bibinfo{person}{Martin Wattenberg}, \bibinfo{person}{Martin Wicke},
  \bibinfo{person}{Yuan Yu}, {and} \bibinfo{person}{Xiaoqiang Zheng}.}
  \bibinfo{year}{2015}\natexlab{}.
\newblock \bibinfo{title}{{TensorFlow}: Large-Scale Machine Learning on
  Heterogeneous Systems}.
\newblock
\newblock
\urldef\tempurl%
\url{https://www.tensorflow.org/}
\showURL{%
\tempurl}
\newblock
\shownote{Software available from tensorflow.org}.


\bibitem[Bannink et~al\mbox{.}(2021)]%
        {bannink2021larq}
\bibfield{author}{\bibinfo{person}{Tom Bannink}, \bibinfo{person}{Arash
  Bakhtiari}, \bibinfo{person}{Adam Hillier}, \bibinfo{person}{Lukas Geiger},
  \bibinfo{person}{Tim de Bruin}, \bibinfo{person}{Leon Overweel},
  \bibinfo{person}{Jelmer Neeven}, {and} \bibinfo{person}{Koen Helwegen}.}
  \bibinfo{year}{2021}\natexlab{}.
\newblock \bibinfo{title}{Larq Compute Engine: Design, Benchmark, and Deploy
  State-of-the-Art Binarized Neural Networks}.
\newblock
\newblock
\showeprint[arxiv]{2011.09398}~[cs.LG]


\bibitem[Chen et~al\mbox{.}(2018)]%
        {scamp2018}
\bibfield{author}{\bibinfo{person}{Jianing Chen}, \bibinfo{person}{Stephen~J
  Carey}, {and} \bibinfo{person}{Piotr Dudek}.}
  \bibinfo{year}{2018}\natexlab{}.
\newblock \showarticletitle{Scamp5d vision system and development framework}.
  In \bibinfo{booktitle}{\emph{Proceedings of the 12th International Conference
  on Distributed Smart Cameras}}. \bibinfo{pages}{1--2}.
\newblock


\bibitem[Gallego et~al\mbox{.}(2020)]%
        {gallego2020event}
\bibfield{author}{\bibinfo{person}{Guillermo Gallego}, \bibinfo{person}{Tobi
  Delbr{\"u}ck}, \bibinfo{person}{Garrick Orchard}, \bibinfo{person}{Chiara
  Bartolozzi}, \bibinfo{person}{Brian Taba}, \bibinfo{person}{Andrea Censi},
  \bibinfo{person}{Stefan Leutenegger}, \bibinfo{person}{Andrew~J Davison},
  \bibinfo{person}{J{\"o}rg Conradt}, \bibinfo{person}{Kostas Daniilidis},
  {et~al\mbox{.}}} \bibinfo{year}{2020}\natexlab{}.
\newblock \showarticletitle{Event-based vision: A survey}.
\newblock \bibinfo{journal}{\emph{IEEE transactions on pattern analysis and
  machine intelligence}} \bibinfo{volume}{44}, \bibinfo{number}{1}
  (\bibinfo{year}{2020}), \bibinfo{pages}{154--180}.
\newblock


\bibitem[Gottardi et~al\mbox{.}(2009)]%
        {sc-sensor}
\bibfield{author}{\bibinfo{person}{Massimo Gottardi}, \bibinfo{person}{Nicola
  Massari}, {and} \bibinfo{person}{Syed~Arsalan Jawed}.}
  \bibinfo{year}{2009}\natexlab{}.
\newblock \showarticletitle{A 100 $\mu$ W 128 $\times$ 64 Pixels Contrast-Based
  Asynchronous Binary Vision Sensor for Sensor Networks Applications}.
\newblock \bibinfo{journal}{\emph{IEEE Journal of Solid-State Circuits}}
  \bibinfo{volume}{44}, \bibinfo{number}{5} (\bibinfo{year}{2009}),
  \bibinfo{pages}{1582--1592}.
\newblock
\urldef\tempurl%
\url{https://doi.org/10.1109/JSSC.2009.2017000}
\showDOI{\tempurl}


\bibitem[Hao(2019)]%
        {AIcarbon2019}
\bibfield{author}{\bibinfo{person}{Karen Hao}.}
  \bibinfo{year}{2019}\natexlab{}.
\newblock \showarticletitle{Training a single AI model can emit as much carbon
  as five cars in their lifetimes}.
\newblock \bibinfo{journal}{\emph{MIT technology Review}}
  (\bibinfo{year}{2019}).
\newblock


\bibitem[Houben et~al\mbox{.}(2013)]%
        {Houben-IJCNN-2013}
\bibfield{author}{\bibinfo{person}{Sebastian Houben}, \bibinfo{person}{Johannes
  Stallkamp}, \bibinfo{person}{Jan Salmen}, \bibinfo{person}{Marc Schlipsing},
  {and} \bibinfo{person}{Christian Igel}.} \bibinfo{year}{2013}\natexlab{}.
\newblock \showarticletitle{Detection of Traffic Signs in Real-World Images:
  The {G}erman {T}raffic {S}ign {D}etection {B}enchmark}. In
  \bibinfo{booktitle}{\emph{International Joint Conference on Neural
  Networks}}.
\newblock


\bibitem[Kubendran et~al\mbox{.}(2021)]%
        {kubendran2021}
\bibfield{author}{\bibinfo{person}{Rajkumar Kubendran}, \bibinfo{person}{Akshay
  Paul}, {and} \bibinfo{person}{Gert Cauwenberghs}.}
  \bibinfo{year}{2021}\natexlab{}.
\newblock \showarticletitle{A 256x256 6.3 pJ/pixel-event Query-driven Dynamic
  Vision Sensor with Energy-conserving Row-parallel Event Scanning}. In
  \bibinfo{booktitle}{\emph{2021 IEEE Custom Integrated Circuits Conference
  (CICC)}}. IEEE, \bibinfo{pages}{1--2}.
\newblock


\bibitem[Li et~al\mbox{.}(2019)]%
        {li2019vehicle}
\bibfield{author}{\bibinfo{person}{Changle Li}, \bibinfo{person}{Quyuan Luo},
  \bibinfo{person}{Guoqiang Mao}, \bibinfo{person}{Min Sheng}, {and}
  \bibinfo{person}{Jiandong Li}.} \bibinfo{year}{2019}\natexlab{}.
\newblock \showarticletitle{Vehicle-mounted base station for connected and
  autonomous vehicles: Opportunities and challenges}.
\newblock \bibinfo{journal}{\emph{IEEE Wireless Communications}}
  \bibinfo{volume}{26}, \bibinfo{number}{4} (\bibinfo{year}{2019}),
  \bibinfo{pages}{30--36}.
\newblock


\bibitem[Patterson et~al\mbox{.}(2021)]%
        {Google2021}
\bibfield{author}{\bibinfo{person}{David Patterson}, \bibinfo{person}{Joseph
  Gonzalez}, \bibinfo{person}{Quoc Le}, \bibinfo{person}{Chen Liang},
  \bibinfo{person}{Lluis-Miquel Munguia}, \bibinfo{person}{Daniel Rothchild},
  \bibinfo{person}{David So}, \bibinfo{person}{Maud Texier}, {and}
  \bibinfo{person}{Jeff Dean}.} \bibinfo{year}{2021}\natexlab{}.
\newblock \showarticletitle{Carbon emissions and large neural network
  training}.
\newblock \bibinfo{journal}{\emph{arXiv preprint arXiv:2104.10350}}
  (\bibinfo{year}{2021}).
\newblock


\bibitem[Sandler et~al\mbox{.}(2019)]%
        {sandler2019mobilenetv2}
\bibfield{author}{\bibinfo{person}{Mark Sandler}, \bibinfo{person}{Andrew
  Howard}, \bibinfo{person}{Menglong Zhu}, \bibinfo{person}{Andrey Zhmoginov},
  {and} \bibinfo{person}{Liang-Chieh Chen}.} \bibinfo{year}{2019}\natexlab{}.
\newblock \bibinfo{title}{MobileNetV2: Inverted Residuals and Linear
  Bottlenecks}.
\newblock
\newblock
\showeprint[arxiv]{1801.04381}~[cs.CV]


\bibitem[Wong et~al\mbox{.}(2018)]%
        {wong2018micronnet}
\bibfield{author}{\bibinfo{person}{Alexander Wong},
  \bibinfo{person}{Mohammad~Javad Shafiee}, {and} \bibinfo{person}{Michael~St.
  Jules}.} \bibinfo{year}{2018}\natexlab{}.
\newblock \bibinfo{title}{MicronNet: A Highly Compact Deep Convolutional Neural
  Network Architecture for Real-time Embedded Traffic Sign Classification}.
\newblock
\newblock
\showeprint[arxiv]{1804.00497}~[cs.CV]


\end{thebibliography}


\end{document}